\documentclass[11pt,a4paper]{article}
\usepackage{times,latexsym}
\usepackage{url}
\usepackage[T1]{fontenc}

\usepackage[acceptedWithA]{tacl2021v1}

\usepackage{xspace,mfirstuc,tabulary}

\newif\iftaclinstructions
\taclinstructionsfalse 
\iftaclinstructions
\renewcommand{\confidential}{}
\renewcommand{\anonsubtext}{(No author info supplied here, for consistency with
TACL-submission anonymization requirements)}
\newcommand{\instr}
\fi

\iftaclpubformat 

\else

\fi

\usepackage[toc,page]{appendix}
\usepackage{adjustbox}
\usepackage{booktabs}
\usepackage{enumitem}

\usepackage{pifont}
\usepackage[nointegrals]{wasysym}
\newcommand{\cmark}{\textcolor{black}{\ding{51}}}%
\newcommand{\xmark}{\textcolor{black}{\ding{55}}}%
\newcommand{\propref}[1]{P\ref*{prop:#1}}

\usepackage{amsmath,amsfonts,amsthm}
\newtheorem{prop}{Property}

\usepackage{comment}
\usepackage{diagbox}
\usepackage[normalem]{ulem}

\usepackage{scalerel,stackengine}
\stackMath
\newcommand\reallywidehat[1]{%
\savestack{\tmpbox}{\stretchto{%
  \scaleto{%
    \scalerel*[\widthof{\ensuremath{#1}}]{\kern-.6pt\bigwedge\kern-.6pt}%
    {\rule[-\textheight/2]{1ex}{\textheight}}
  }{\textheight}%
}{0.5ex}}%
\stackon[1pt]{#1}{\tmpbox}%
}

\newcommand*\rot{\rotatebox{90}}

\title{A Closer Look at Classification Evaluation Metrics and a Critical Reflection of Common Evaluation Practice}

\author{
  Juri Opitz
  \\
  University of Zurich
  \\
  \texttt{opitz.sci@gmail.com}
}

\date{}

\begin{document}
\maketitle
\begin{abstract}
Classification systems are evaluated in a countless number of papers. However, we find that evaluation practice is often nebulous. Frequently, metrics are selected without arguments, and blurry terminology invites misconceptions. For instance, many works use so-called `macro' metrics to rank systems (e.g., `macro F1') but do not clearly specify what they would expect from such a `macro' metric. This is problematic, since picking a metric can affect research findings, and thus any clarity in the process should be maximized. 

Starting from the intuitive concepts of \textit{bias} and \textit{prevalence}, we perform an analysis of common evaluation metrics. The analysis helps us understand the metrics' underlying properties, and how they align with expectations as found expressed in papers. Then we reflect on the practical situation in the field, and survey evaluation practice in recent shared tasks. We find that metric selection is often not supported with convincing arguments, an issue that can make a system ranking seem arbitrary. Our work aims at providing overview and guidance for more informed and transparent metric selection, fostering meaningful evaluation.  
\end{abstract}

\section{Introduction}

Classification evaluation is ubiquitous: we have a system that predicts some classes of interest (aka classifier) and want to assess its prediction skill. We study a widespread and seemingly clear-cut setup of multi-class evaluation, where we compare a classifier's predictions against reference labels in two steps. First, we construct a \textit{confusion matrix} that has a designated dimension for each possible prediction/label combination. Second, an aggregate statistic, which we denote as \textit{metric}, summarizes the confusion matrix as a single number. 

Already from this it would follow that `the perfect' metric can't exist, since important information is bound to get lost when reducing the confusion matrix to a single dimension. Still, we require a metric to rank and select classifiers, and thus it should characterize a classifier's `skill' or `performance' as well as possible. But exactly what is a `performance' and how should we measure it? Such questions do not seem to arise (as much) in other `performance' measurements that are known to humans. E.g., a marathon's result derives from a clear and broadly accepted criterion (time over distance) that can be measured with validated instruments (a clock). However, in machine learning, criterion and instrument are often less clear and lie entangled in the term `metric'.

Since metric selection can influence which system we consider better or worse in a task, one would think that metrics are selected with great care. But when searching through papers for reasons that would support a particular metric choice, we mostly (at best) find only weak or indirect arguments. E.g., it is observed (Table \ref{tab:practices}\footnote{Example excerpts are taken from  \citet{ van-hee-etal-2018-semeval, barbieri-etal-2018-semeval, zampieri-etal-2019-semeval, dimitrov-etal-2021-semeval, ding-etal-2020-discriminatively, yi-etal-2019-towards, xing-etal-2020-financial, NEURIPS2021_bcd0049c-accuracy-macro, wu-etal-2023-dont}.}) that `labels are imbalanced' or it is wished for that `all class labels have equal weight'. These perceived problems or needs are then often supposedly addressed with a `macro' metric (Table \ref{tab:practices}). 

However, what is meant with phrases like `imbalanced data' or `macro' is rarely made explicit, and how the metrics that are then selected in this context are actually addressing a perceived `imbalance' is unclear. According to a word etymology, evaluation with a `macro' metric may involve the expectation that we are told a \textit{bigger picture} of classifier capability (Greek: makrós, `long/large'), whereas a smaller picture (Greek: mikrós, `small') would perhaps bind the assessment to a more local context. Regardless of such musings, it is clear that blurry terminology in the context of classifier evaluation can lead to results that may be misconceived by readers and authors alike. 

\begin{table}
    \centering
    \scalebox{0.701}{
    \begin{tabular}{ll}
    \toprule
        metric & cited motivation/argument \\
        \midrule
        macro F1           & ``macro-averaging (...) implies that all class \\ 
                           &~~labels have equal weight in the final score''\\
        macro P/R/F1       & ``because (...) skewed distribution \\ 
                           &~~of the label set'' \\
        macro F1           & ``Given the strong imbalance between the \\ 
                           &~~number of  instances in the different classes'' \\
        Accuracy, macro F1 & ``the labels are imbalanced'' \\
        MCC                &  ``balanced measurement when the classes \\ 
                           &~~are of very different sizes''\\
        MCC, F1            & ``(...) imbalanced data (...)'' \\ 
        macro F1           & ``(...) imbalanced classes (...) introduce \\ 
                           &~~biases on accuracy'' \\
        macro F1           & ``due to the imbalanced dataset''\\
    \bottomrule
    \end{tabular}}
    \caption{Typical comments that intend to motivate metric selection. MCC: \textit{Matthews Correlation Coefficient}.}
    \label{tab:practices}
\end{table}

This paper aims to serve as a handy reference for anyone who wishes to better understand classification evaluation, how evaluation metrics align with expectations expressed in papers, and how we might construct an individual case for (or against) selecting a particular evaluation metric. 

\paragraph{Paper outline.} After introducing \hyperref[sec:prelim]{\textit{Preliminaries} (\S \ref{sec:prelim})} and five \hyperref[sec:properties]{\textit{Metric Properties} (\S \ref{sec:properties})}, we conduct a thorough \hyperref[sec:overview]{\textit{Metric Analysis} (\S \ref{sec:overview})} of common classification measures: `Accuracy' (\S \ref{subsec:accuracy}), `Macro Recall and Precision' (\S \ref{subsec:macR}, \S \ref{subsec:macP}), two  different `Macro F1' (\S \ref{subsec:macF}, \S \ref{subsec:macFbar}), `Weighted F1' (\S \ref{subsec:wf1}), as well as `Kappa' and `Matthews Correlation Coefficient (MCC)' in \S \ref{subsec:kappamcc}. We then show how to create simple but meaningful \hyperref[sec:variants]{\textit{Metric Variants} (\S \ref{sec:variants})}.  We wrap up the theoretical part with a \hyperref[sec:discussion]{\textit{Discussion} (\S \ref{sec:discussion})} that includes a short \hyperref[sec:summary]{\textit{Summary} (\S \ref{sec:summary})} of our main analysis results in \hyperref[tab:summary]{Table \ref{tab:summary}}. Next we study \hyperref[sec:semeval]{\textit{Metric Selection in Shared Tasks}} (\S \ref{sec:semeval}) and give \hyperref[sec:recommendations]{\textit{Recommendations}} (\S \ref{sec:recommendations}). Finally, we contextualize our work against some \hyperref[sec:rw]{\textit{Background and Related Work} (\S \ref{sec:rw})}, and finish with \hyperref[sec:remarks]{\textit{Conclusions}} (\S \ref{sec:remarks}).

\section{Preliminaries}
\label{sec:prelim}

We introduce a set of intuitive concepts as a basis.

\paragraph{\textit{Classifier}, \textit{confusion matrix}, and \textit{metric}.} For any classifier  $f : D \rightarrow C=\{1,...,n\}$ and finite set $S \subseteq D\times C $, let  $m^{f,S}\in\mathbb{R}_{\geq 0}^{n\times n}$ be a confusion matrix where $m^{f,S}_{ij}=|\{s\in S~|~f(s_1)=i \land  s_2=j\}|$.\footnote{We use $\mathbb{R}$ (instead of $\mathbb{N}$) to allow for cases where matrix fields contain, e.g., ratios, or accumulated `soft' scores.} 
We omit superscripts whenever possible. So generally, $m_{i, j}$ contains the mass of events where the classifier predicts $i$, and the true label is $j$. That is, on the diagonal of the matrix lies the mass of correct predictions, and all other fields indicate the mass of specific errors. A $metric: \mathbb{R}_{\geq 0}^{n\times n} \rightarrow (-\infty, 1]$ then allows us to order confusion matrices, respectively, rank classifiers (bounds are chosen for convenience). We say that (for a data set $S$) a classifier $f$ is better than (or preferable to) a classifier $g$ iff $metric(m^{f, S}) > metric(m^{g, S})$, i.e., a higher score indicates a better classifier.

Let us now define five basic quantities:

\paragraph{Class \textit{bias}, \textit{prevalence} and \textit{correct}} are given as
\begin{align*}
    bias(i) = \sum_{x}m_{i,x}~~~~~~ prevalence(i) = \sum_{x} m_{x,i} \\ correct(i) = m_{i,i}.~~~~~~~~~~~~~~~~~~~~~~~~~
\end{align*}

\paragraph{Class precision.} $P_i$ is the precision for class $i$: 
\begin{equation}
  P_i = \frac{correct(i)}{bias(i)} \approx \mathcal{P}(c=i|f \rightarrow i).
\end{equation}
It approximates the probability of observing a correct class given a specific prediction.

\paragraph{Class recall.} $R_i$ denotes the recall for class $i$,
\begin{equation}
 R_i= \frac{correct(i)}{prevalence(i)} \approx \mathcal{P}(f \rightarrow i |c=i), 
\end{equation}
that approximates the probability of observing a correct prediction given an input of a certain class.

\section{Defining metric properties}
\label{sec:properties}

To understand and distinguish metrics in more precise ways, we define five metric properties: \textit{Monotonicity}, \textit{class sensitivity}, \textit{class decomposability}, \textit{prevalence invariance} and \textit{chance correction}.

\subsection{Monotonicity (PI)}

Take a classifier that receives an input. If the prediction is correct, we would naturally expect that the evaluation score does not decrease, and if it is wrong, the evaluation score should not increase. We cast this clear expectation into
\begin{prop}[Monotonicity] \label{prop:mo}
A $metric$ has \propref{mo} iff:
\begin{equation}
\forall m: \frac{\partial metric(m)}{\partial m_{i,j}}  \begin{cases}
\geq 0 \impliedby i = j\\
\leq 0 \impliedby i \neq j,
\end{cases}
\end{equation}
\end{prop}
\noindent i.e., diagonal fields of the confusion matrix (correct mass) should yield a non-negative `gradient' in the metric, while for all other fields (containing error mass) it should be non-positive. \propref{mo} assumes differentiability of a metric, but it can be simply extended to the discrete case.\footnote{Assume any data set $S$ and split $S', S'', S'''$ s.t.\ $S' \cup S'' \cup S''' = S$ and $|S'' = \{(x, y)\}| = 1$. Then for any classifier $f$ we want to ensure $f(x) = y \implies metric(m^{f, S' \cup S''}) \geq metric(m^{f, S'})$, else $metric(m^{f, S' \cup S''}) \leq metric(m^{f, S'})$.}

\subsection{Macro metrics are class-sensitive (PII)}

A `macro' metric needs to be sensitive to classes, or else it could not yield a `balanced measurement' for `classes having different sizes' (c.f.\ Table \ref{tab:practices}). By contrast, a `micro' metric should care only about whether predictions are wrong or right, which would bind its score more to a local context of a specific data set and its class distribution. This means for macro metrics that they should possess 

\begin{prop}[Class sensitivity]\label{prop:cs}
\propref{cs} is given iff $\exists m \in \mathbb{R}_{\geq 0}^{|C| \times |C|}$ with $ \frac{\partial metric(m)}{\partial m_{i,i}} \neq \frac{\partial metric(m)}{\partial m_{j,j}} $, $(i,j) \in C^2$ or $\frac{\partial metric(m)}{\partial m_{i,j}} \neq \frac{\partial metric(m)}{\partial m_{k,l}}$, $(i, j, k, l) \in C^4, i \neq j, k \neq l$.\footnote{Discrete case: Assume any data set $S$ and split $S', S'', S''', S''''$ s.t.\ $S' \cup S'' \cup S''' \cup S'''' = S$ and $|S'' = \{(x, y)\}| = |S ''' = \{(w, z)\}| = 1$. Then $metric$ is not a `micro' metric if there is any $f$ with $[f(x) = y \land f(w) = z] \lor [f(x) \neq y \land f(w) \neq z] $ and $ metric(m^{f, S' \cup S''}) \neq metric(m^{f, S' \cup S'''})$.} 
\end{prop}

\noindent A metric without \propref{cs} is not a macro metric.

\subsection{Macro average: Mean over classes (PIII)}

`Macro' metrics are sometimes named `macro-average' metrics. This suggests that they may be perceived as an average over classes. We introduce 

\begin{prop}[Class decomposability] \label{prop:cd} A `macro-average' metric can be stated as 
\begin{equation}
\label{eq:mean}
    metric(m)^g_p= \left(\frac{1}{n} \sum_{i=1}^n g(m_{i}, (m^T)_{i}, i)^p \right)^{\frac{1}{p}},
\end{equation}
\end{prop}
\noindent i.e., as an unweighted mean over class-specific scores from inputs examples related to a specific class ($c=i$ $\lor$ $f\rightarrow i$). E.g., later we will see that  `macro F1' is a specific parameterization of Eq.\ \ref{eq:mean}.

\subsection{Strictly ``Treat all classes equally'' (PIV)} A common argument for using metrics other than the ratio of correct predictions is that we want to `not neglect \textit{rare} classes' and `show classifier performance \textit{equally} [w.r.t.] all classes'.\footnote{See also Table \ref{tab:practices}, and, e.g., \citet{m4,m1,m2}.}

Nicely, with the assumption of class prevalence being the only important difference across data sets, we could then even say that classifier $f$ is \textit{generally} better than (or preferable to) $g$ iff $metric(m^{f}) > metric(m^{g})$, further disentangling a classifier comparison from a specific data set. At first glance, \propref{cd} seems to capture this wish already, by virtue of an \textit{unweighted mean} over classes. However, the score w.r.t.\ one class is still influenced by the prevalence of other classes, and thus the result of the mean can change in non-transparent ways if class frequency is varied.
\begin{table*}
\begin{minipage}{.32\linewidth}
    \centering
    \begin{tabular}{p{0.2cm}ccc}
          & & \multicolumn{2}{c}{$c=$} \\
          & & \multicolumn{1}{|c}{x} & y \\
        \cmidrule{2-4}
       & x  & \multicolumn{1}{|c}{15} & 5 \\
       \rot{\rlap{$f$$\rightarrow$}} 
       & y   & \multicolumn{1}{|c}{10} & 10 \\
    \end{tabular}
    \caption{Class $y$ occurs 15 times.}
    \label{tab:subtab1}
\end{minipage}%
\begin{minipage}{.32\linewidth}
    \centering
    \begin{tabular}{p{0.2cm}ccc}
          & & \multicolumn{2}{c}{$c=$} \\
          & & \multicolumn{1}{|c}{x} & y \\
        \cmidrule{2-4}
       & x   & \multicolumn{1}{|c}{15 $\cdot 1$} & 5 $\cdot 2$ \\
       \rot{\rlap{$f$$\rightarrow$}} 
       & y   & \multicolumn{1}{|c}{10 $\cdot 1$} & 10 $\cdot 2$\\
    \end{tabular}
    \caption{Apply $\mathbb{\lambda}=(1,2)$.}
    \label{tab:subtab2}
\end{minipage}
\begin{minipage}{.32\linewidth}
    \centering
    \begin{tabular}{p{0.2cm}ccc}
          & & \multicolumn{2}{c}{$c=$} \\
          & & \multicolumn{1}{|c}{x} & y \\
        \cmidrule{2-4}
       & x   & \multicolumn{1}{|c}{15} & 10 \\
       \rot{\rlap{$f$$\rightarrow$}}
       & y   & \multicolumn{1}{|c}{10} & 20 \\
    \end{tabular}
    \caption{Class $y$ occurs 30 times.}
    \label{tab:subtab3}
\end{minipage}
\label{tab:ex1}
\end{table*}

Therefore it makes sense to define such an expectation (`treat all classes equally') more strictly. We simulate different class prevalences with a

\paragraph{Prevalence scaling.} We can use a diagonal prevalence scaling matrix $\mathbb{\lambda}$ to set
\begin{equation}
\label{eq:transf}
    m' = m\mathbb{\lambda}.
\end{equation}
By scaling a column $i$ with $\lambda_{ii}$, we inflate (or deflate) the mass of data that belong to class $i$ (e.g., see Tables \ref{tab:subtab1}, \ref{tab:subtab2}, \ref{tab:subtab3}), but retain the relative proportions of intra-class error types. Now, we can define

\begin{prop}[Prevalence invariance]\label{prop:pi} If $(\mathbb{\lambda}, \mathbb{\lambda}') \in \mathbb{R}_{>0}^{n\times n} \times \mathbb{R}_{>0}^{n\times n}$ is a pair of diagonal matrices then $metric(m\mathbb{\lambda}) = metric(m\mathbb{\lambda}')$.
\end{prop}

\paragraph{Prevalence calibration.} There is a special case of $\lambda$. We select $\lambda^\sim$ s.t.\ all classes have the same prevalence. We call this prevalence calibration:
\begin{equation}
\label{eq:calibrate}
    \lambda_{ii}^\sim = \frac{|S|}{n\cdot prevalence(i)}.
\end{equation}

\subsection{Chance correction (PV)} 

Two simple `baseline' classifiers are: Predicting classes uniformly randomly, or based on observed prevalence. A macro metric can be expected to show robustness against \textit{any} such chance classifier and be \textit{chance corrected}, assigning a clear and comparable baseline score. Thus, it should have
\begin{prop}[Chance Correction]\label{prop:cc} A $metric$ has \propref{cc} iff for any (large) dataset $S$ with $n$ classes and set $A$ with arbitrary random classifiers: 
\begin{align*}
\max \big\{ metric(m^{r,S})~~|~~ r \in A\big\} = \omega(n^S) << 1. 
    \end{align*}
\end{prop}
\noindent Here, $\omega$ returns an upper-bound baseline score from the number of classes $n^S$ alone. If it also holds that $\max \{ ...\} = \min \{...\} =\omega(n^S)$, we say that $metric$ is \textit{strictly chance corrected}, and in the case where $\forall S, r:~metric(m^{r,S}) = \Omega~\text{(constant)}$ we speak of \textit{complete chance correction}. 

Less formally, chance correction means that the metric score attached to any chance baseline has a bound that is known to us (the bound generalizes over data sets but not over the number of classes). Strict chance correction means additionally that any chance classifier's score will be the same, and just depends on the number of classes. Finally, complete chance correction means that every chance classifier always yields the same score, regardless of the number of classes. Note that strictness or completeness may not always be desired, since they can marginalize empirical overall correctness in a data set. Any chance correction, however, increases the evaluation interpretability by contextualizing the evaluation with an interpretable baseline score.

\section{Metric property analysis}
\label{sec:overview}

Equipped with the appropriate tools, we are now ready to start the analysis of classification metrics. We will study `Accuracy', `Macro Recall', `Macro Precision', `Macro F1', `Weighted F1', `Kappa', and `Matthews Correlation Coefficient' (MCC).

\subsection{Accuracy (aka Micro F1)}\label{subsec:accuracy}

Accuracy is the ratio of correct predictions:
\begin{equation*}
\label{eq:acc}
    accuracy = \frac{\sum_{i}m_{i,i}}{\sum_{(i, j)}m_{i,j}} = \frac{1}{|S|}\sum_{i} correct(i).
\end{equation*} 

\paragraph{Property analysis.}  As a `micro' metric, Accuracy has only \propref{mo} (monotonicity). This is expected, since PII-V aim at \textit{macro} metrics. Interestingly, in multi-class evaluation, $accuracy$ equals `\textit{micro Precion, micro Recall and micro F1}' that sometimes occur in papers. See Appendix \ref{app:micacc} for proofs.

\paragraph{Discussion.} Accuracy is an important statistic, estimating the probability of observing a correct prediction in a data set. But this means that it is strictly tied to the class prevalences in a specific data set. And so, in the pursuit of some balance or a more generalizable score, researchers seem interested in other metrics.

\subsection{Macro Recall: ticks five boxes}\label{subsec:macR}

Macro Recall is defined as the unweighted arithmetic mean over all class-wise recall scores:
\begin{equation}
    macR = \frac{1}{n}\sum_i R_i = \frac{1}{n}\sum_i \frac{correct(i)}{prevalence(i)}.
\end{equation}

\paragraph{Property analysis.} Macro Recall has all five properties (Proofs in Appendix \ref{app:anamacr}). It is also \textit{strictly} chance corrected with $\omega(n) = 1/n$.

\paragraph{Discussion.} Since macro Recall has all five properties, including prevalence invariance (\propref{pi}), it may be a good pick for evaluation, particularly through a `macro' lens. It also offers three intuitive interpretations: \textit{Drawing an item from a random class}, \textit{Bookmaker metric} and \textit{prevalence-calibrated Accuracy}. 

In the first interpretation, we draw a random item from a randomly selected class. What's the probability that it is correctly predicted? $MacR$ estimates the answer $\sum_i\frac{1}{n}\cdot \mathcal{P}(f\rightarrow i | c=i)$.

Alternatively, we wear the lens of \textit{a (fair) Bookmaker}.\footnote{On bookmaker inspired metrics cf.\ \citet{powers2003recall, powers2020evaluation}.} For every prediction (bet), we pay 1 coin and gain coins per fair (European) odds. The odds for making a correct bet, when the class is $i$, are $odds(i) = \frac{|S|}{prevalence(i)}$. So for each data example $(x, y)$, our bet is evaluated ($\mathbb{I}[f(x) = y] \in \{0,1\}$), and thus we incur a total net 
\begin{align*}
    win &= \sum_{s \in S}  \bigg(\mathbb{I}[f(s_1) = s_2]\cdot odds(s_2)- 1\bigg) \\
    &= \sum_{s \in S}   \bigg(\mathbb{I}[f(s_1) = s_2]\cdot odds(s_2)\bigg) -|S| \\  
    &= \sum_{i=1}^n \bigg[ odds(i) \cdot \sum_{\substack{s \in S \\ s_2=i}} \bigg(\mathbb{I}[f(s_1) = i]\bigg)\bigg] -|S|\\
    &= |S|\sum_{i=1}^{n} \frac{correct(i)}{prevalence(i)} -|S| \\ &=  n|S|\cdot macR -|S| = |S|(n \cdot macR - 1),
\end{align*}
which is positive only if $macR > 1/n$.

Finally we can view \textit{macro Recall as Accuracy after prevalence calibration}. Set $\lambda$ as in Eq.\ \ref{eq:calibrate}:
\begin{align}
\label{eq:macacc=macrSTART}
macAcc &= accuracy(m\lambda^\sim)\nonumber\\ 
&= \frac{\sum_i{\lambda_{ii}^\sim\cdot correct(i)}}{\sum_i{\lambda_{ii}^\sim\cdot prevalence(i)}} \\ 
&=\sum_{i} R_i/\sum_{i} 1 = macR.\nonumber
\label{eq:macacc=macrEND}
\end{align}

\subsection{Macro Precision: is the bias an issue?}\label{subsec:macP}

Macro Precision is the unweighted arithmetic mean over class-wise precision scores:
\begin{equation}
    macP = \frac{1}{n}\sum_i P_i = \frac{1}{n}\sum_i \frac{correct(i)}{bias(i)}.
\end{equation}

\paragraph{Property analysis.} While properties I, II, III, V are fulfilled, macro Precision does not have prevalence invariance (Proofs in Appendix \ref{app:anamacp}). 
With some $\lambda$, the max.\ score difference ($macP(m)$ vs.\ $macP( m\lambda)$) approaches $1 - \frac{1}{n}$.\footnote{Consider a matrix with ones on the diagonal, and large numbers in the first column (yielding low class-wise precision scores). With $\lambda$ where $\lambda_{1,1}$ is very small (reducing the prevalence of class $1$), we obtain high precision scores.} Like macro Recall, it is strictly chance corrected ($\omega(n)=1/n$).

\paragraph{Discussion.} Macro Precision wants to approximate the probability to see a correct prediction, given we randomly draw one out of $n$ different predictions. Hence, $macP$ seems to provide us with an interesting measure of `prediction trustworthiness'. An issue is that the score does not generalize across different class prevalences, since $bias(i) \propto \mathcal{P}(f \rightarrow i) =  \sum_j \mathcal{P}(f \rightarrow i, c=j)$ is subject to change if prevalences of other classes $j\neq i$ vary ($\propto$ means approximately proportional to). Therefore, even though $macP$ is decomposed over classes (\propref{cd}), it is not invariant to prevalence changes (\propref{pi}), and if we have $f, f'$ with different biases, score differences are difficult to interpret, particularly with an underlying `macro' expectation that a metric be robust to class prevalence.

To mitigate the issue, we can use prevalence calibration (Eq.\ \ref{eq:calibrate}), yielding
\begin{align*}
&correct^\sim(i) = \lambda_{i,i}^\sim m_{i, i} \propto \mathcal{P}(f \rightarrow i | c= i),\\
&bias^\sim(i) = \sum_j  \lambda_{j,j}^\sim m_{i,j} \propto \sum_j \mathcal{P}(f\rightarrow i | c=j), 
\end{align*}
and a $macP^\sim$ that employs a prior belief that all classes have the same prevalence. Like macro Recall, $macP^\sim$ is now detached from the class distribution in a specific data set, treating all classes more literally `equally'.

\subsection{Macro F1: Metric of choice in many tasks}\label{subsec:macF}
\label{subsec:macrof1}

Macro F1 is often used for evaluation. It is commonly defined as an arithmetic mean over class-wise harmonic means of precision and recall: 
\begin{align}
\label{eq:macf1normal}
macF1 &= \frac{1}{n}\sum_{i}F1_i = \frac{1}{n}\sum_{i} \frac{2P_iR_i}{P_i+ R_i}\\
&= \frac{2}{n}\sum_{i}  \frac{correct(i)}{bias(i) + prevalence(i)}.\nonumber
\end{align}
\paragraph{Property analysis.} Again, all properties except \propref{pi} are fulfilled (Proofs in Appendix \ref{app:anamacf1}). Interestingly, while macro F1 has \propref{cc} (chance correction), the chance correction isn't strict, differentiating it from other macro metrics: Indeed, its chance baseline upper bound $\omega(n)=1/n$ is achieved only if $\mathcal{P}(f \rightarrow i) = \mathcal{P}(c=i)$, meaning that macro F1 not only corrects for chance, but also factors in more data set accuracy (like a `micro' score). Additionally, the second line of the formula shows that macro F1 is invariant to the false-positive and false-negative error spread for a specific class.

\paragraph{Discussion.} Macro F1 wants the distribution of prediction and class prevalence to be similar (a micro feature), but also high correctness for every class, by virtue of the unweighted mean over classes (a macro feature). Thus it seems useful to find classifiers that do well in a given data set, but probably also in others, a `balance' that could explain its popularity. However, macro F1 inherits an interpretability issue of Precision. It doesn't strictly `treat all classes equally' as per \propref{pi}, at least not without prevalence calibration (Eq.\ \ref{eq:calibrate}).

\subsection{Macro F1: A doppelganger}\label{subsec:macFbar}

Interestingly, there is another metric that has been coined `macro F1'. We find an early mention in \citet{sokolova2009systematic} and evaluation usage (made explicit) in a lot of papers, i.a., \citet{stab-gurevych-2017-parsing, mohammadi-etal-2020-cooking, rodrigues-branco-2022-transferring}. This macro F1 is the harmonic mean of macro Precision and Recall:
\begin{equation}
\label{eq:macf1bar}
    macF1' = \frac{2 \cdot macR \cdot macP}{macR + macP}.
\end{equation}

\paragraph{Property analysis.} In contrast to its name twin, one less property is given (\propref{cd}), since it cannot be decomposed over classes (Proofs in Appendix \ref{app:anamacf1bar}), and it is \textit{strictly} chance corrected with $\omega(n)=1/n$. \citet{opitz2019macro} prove that Eq.\ \ref{eq:macf1bar} and Eq.\ \ref{eq:macf1normal} can diverge to a large degree of up to 0.5.

\paragraph{Discussion.} Putting the harmonic mean on the outside, and the arithmetic means on the inside, $macF1'$ seems to stick a tad more true to the emphasis in its name (F1, aka harmonic mean). However, $macF1'$ does not seem as easy to interpret, since the numerator involves the cross-product of all class-wise recall and precision values. We might view it through the lens of an inter-annotator agreement (IAA) metric though, treating classifier and reference as two annotators:
\begin{equation}
    macF1' = \frac{2 \cdot macR(m) \cdot macR(m^T)}{macR(m) + macR(m^T)},
\end{equation}
falling back on $macR$'s clear interpretation(s).

\subsection{Weighted F1}
\label{subsec:wf1}

`Weighted F1' or `Weighted average F1' is yet another F1 variant that has been used for evaluation:
\begin{align*}
    &weightF1 = \frac{1}{|S|} \sum_{i} prevalence(i) \cdot F1_i.
\end{align*}
\paragraph{Property analysis.} Weighted F1 is sensible to classes (\propref{cs}). The other four properties are not featured, which means that it is also non-monotonic. See Appendix \ref{app:wf1} for proofs.

\paragraph{Discussion.} While measuring performance `locally' for each class, the results are weighted by class-prevalence. Imagining metrics on a spectrum from `micro' to `macro', $weightF1$ sits next to Accuracy, the prototypical micro metric. This is also made obvious by its featured properties, where only one would mark a `macro' metric (\propref{cs}). Due to its lowered interpretability and non-monotonicity, we may wonder why $weightF1$ would be preferred over Accuracy. Finally, with prevalence calibration, it reduces to macro F1, $weightF1(m\lambda^\sim) = macF1(m\lambda^\sim)$, similar to how calibrated Accuracy reduces to macro Recall.

\subsection{Birds of a feather: Kappa and MCC} \label{subsec:kappamcc}

Assuming normalized confusion matrices\footnote{$m_{ij}=\frac{1}{|S|}|\{s\in S~|~f(s_1)=i \land  s_2=j\}| \in [0,1], s.t.\ \sum_{(i,j)}m = 1$. This models ratios in the matrix fields but does not change $MCC$ or $KAPPA$.}, we can state both metrics as concise as possible. Let $\mathbf{1}$ be a vector with ones of dimension $n$. Then let
\begin{align}
\label{eq:kappa_mcc_helper_def}
\mathbf{b}=m\mathbf{1}; \text{~~}
\mathbf{p}=m^T\mathbf{1};  \text{~~} chance = \mathbf{p}^T\mathbf{b}.
\end{align}
I.e., at index $i$ of vector $\mathbf{p}$, we find $prevalence(i)$, and at index $i$ of vector $\mathbf{b}$ we find $bias(i)$. 

\paragraph{Generalized Matthews correlation coefficient (MCC).} The multi-class generalization of MCC \cite{gorodkin2004comparing} can now be written concisely as
\begin{equation}
    \label{eq:mcc}
    MCC = \frac{accuracy -chance}{(\sqrt{1 - \mathbf{b}^T\mathbf{b}})(\sqrt{1 - \mathbf{p}^T\mathbf{p}})}.
\end{equation}

\paragraph{Cohen's kappa} \cite{cohen1960coefficient} is then denoted as follows, illuminating its similarity to MCC:
\begin{equation}
    \label{eq:kappa}
    KAPPA  = \frac{accuracy - chance}{1 - chance}.
\end{equation}

\paragraph{Property analysis.} MCC and Kappa have \propref{cs} and \propref{cc} (complete chance correction: $\Omega=0$). However, they are \textit{non-}monotonic (\propref{mo}), not class-decomposable (\propref{cd}), and not prevalence-invariant (\propref{pi}); Proofs in Appendix \ref{app:anamcckappa}. Further note that $sgn(MCC)=sgn(KAPPA)$ and $|MCC| \geq |KAPPA|$, since $\mathbf{p}^T\mathbf{b}\leq \{\mathbf{b}^T\mathbf{b}, \mathbf{p}^T\mathbf{p}\}$. 

\paragraph{Discussion.} Kappa and MCC are similar measures. Since $chance \approx \sum_{i}\mathcal{P}(c=i)\cdot\mathcal{P}(f \rightarrow i)$ allows the interpretation of observing a prediction that is correct just by chance, Kappa and MCC can be viewed as a standardized Accuracy. 

However, overall they are standardized in slightly different ways. The denominator of Kappa simply shows the upper bound, i.e., the perfect classifier, which is intuitive. How do we interpret $\mathbf{b}^T\mathbf{b}$ and $\mathbf{p}^T\mathbf{p}$ in MCC? Given two random items drawn from two random classes, $\mathbf{b}^T\mathbf{b}$ seems to measure the chance that the classifier randomly predicts the same label, while $\mathbf{p}^T\mathbf{p}$ measures the chance that the true labels are the same. This adds complexity to the MCC formula that can make classifier comparison less clear. The stronger dependence on \textit{classifier bias} through $\mathbf{b}^T\mathbf{b}$ also favors classifiers with uneven biases, regardless of the actual class distribution in a data set. This reduced interpretability is still evident when the measures are prevalence-calibrated (Eq.\ \ref{eq:calibrate}):
\begin{align}
    \label{eq:kappa_calib}
    &KAPPA(m\lambda^\sim)  = \frac{macR - 1/n}{1 - 1/n} \\
    &MCC(m\lambda^\sim)  = \frac{macR - 1/n}{(\sqrt{1 - \mathbf{b^\sim}^T\mathbf{b^\sim}})(\sqrt{1 - 1/n})} \nonumber,
\end{align}
which reduces KAPPA (but not MCC) to $macR$.

\begin{table*}
    \centering
    \scalebox{0.85}{
    \begin{tabular}{l|lllll}
    \toprule
       metric  &  PI (mono.) & PII (class sens.) & PIII (decomp.) & PIV (prev.\ invar.) & PV (chance correct.) \\
       \midrule
       Accuracy (=Micro F1)  & \cmark & \xmark$_\text{(\cmark)}$ & \xmark$_\text{(\cmark)}$ & \xmark$_\text{(\cmark)}$ & \xmark$_\text{(\cmark)}$\\
       macro Recall ($macR$)   & \cmark & \cmark & \cmark & \cmark & \cmark: $1/n$, strict\\
       ~~~as $GmacR$ or $HmacR$ & \cmark & \cmark & \cmark & \cmark & \cmark: $1/n$\\
       macro Precision   & \cmark& \cmark & \cmark & \xmark$_\text{(\cmark)}$ & \cmark: $1/n$, strict\\
       macro F1 ($macF1$)  & \cmark& \cmark & \cmark & \xmark$_\text{(\cmark)}$ &\cmark: $1/n$ \\
       macro F1' ($macF1'$)  & \cmark & \cmark &  \xmark & \xmark$_\text{(\cmark)}$ &\cmark: $1/n$, strict\\
       weighted F1 & \xmark & \cmark &  \xmark$_\text{(\cmark)}$ & \xmark$_\text{(\cmark)}$ &\xmark$_\text{(\cmark)}$ \\ 
       Kappa  & \xmark & \cmark & \xmark & \xmark$_\text{(\cmark)}$& \cmark: 0, complete\\
       MCC  & \xmark & \cmark & \xmark & \xmark$_\text{(\cmark)}$ &\cmark: 0, complete\\
       \bottomrule
    \end{tabular}}
    \caption{Summary of evaluation metric properties. \xmark$_\text{(\cmark)}$: a property is fulfilled after prevalence calibration.}
    \label{tab:summary}
\end{table*}

\section{Metric variants}
\label{sec:variants}

\subsection{Mean parameterization in \propref{cd}} It is interesting to consider $p\neq 1$ in the generalized mean (Eq.\ \ref{eq:mean}). 
E.g., in the example of macro Recall, setting $p \rightarrow 0$ yields the geometric mean
\begin{equation*}
    GmacR = GM(R_1,..., R_n) = \sqrt[n]{R_1 \cdot ... \cdot R_n}.
\end{equation*}
Same as $macR$, it has all five properties. Given $n$ random items, one from every class, $GmacR$ approximates a (class-count normalized) probability that all are correctly predicted. Hence, $GmacR$ can be useful when it's important to perform well in \textit{all} classes. Thinking further along this line, we can employ $HmacR$ ($p = -1$) with the harmonic mean $HM(R_1,..., R_n) = n(\frac{1}{R_1} + ... + \frac{1}{R_n})^{-1}$.

\subsection{Prevalence calibration}
\label{subsec:cmcalib}

Property \propref{pi} (prevalence invariance) is rare, but we saw that it can be artificially enforced. Indeed, if we standardize the confusion matrix by making sure every class has the same prevalence (Eq.\ \ref{eq:transf}, Eq. \ref{eq:calibrate}), we ensure prevalence invariance (\propref{pi}) for a measure. As an effect of this, we found that Kappa and Accuracy reduce to macro Recall, and weighted F1 becomes the same as macro F1. For a more detailed interpretation of prevalence calibration, see our discussion for macro Precision (\S \ref{subsec:macP}).

When does a prevalence calibration make sense? Since prevalence calibration offers a gain in `macro'-features, it can be used with the aim to push a metric more towards a `macro' metric.

\section{Discussion}
\label{sec:discussion}

\subsection{Summary of metric analyses}
\label{sec:summary}

Table \ref{tab:summary} shows an overview of the visited metrics. We make some observations: i) macro Recall has all five properties, including class prevalence invariance (\propref{pi}), i.e., `it treats all classes equally' (in a strict sense). However, through prevalence calibration, all metrics obtain  \propref{pi}. ii) Kappa,  MCC, and weighted F1 do not have property \propref{mo}. Under some circumstances, errors can increase the score, possibly lowering interpretability. iii) All metrics except Accuracy and weighted F1 show chance baseline correction. Strict chance baseline correction isn't a feature of Macro F1, and complete (class-count independent) chance correction is only achieved with MCC and Kappa.

Macro Recall and Accuracy seem to complement each other. Both have a clear interpretation, and relate to each other with a simple prevalence calibration. Indeed, macro Recall can be understood as a prevalence-calibrated version of Accuracy. On the other hand, macro F1 is interesting since it does not strictly correct for chance (as in macro Recall) but also factors in more of the test set correctness (as in a `micro' score). 

MCC and Kappa are similar measures, where Kappa tends to be slightly more interpretable and shows more robustness to classifier biases. Somewhat also similar are Accuracy and weighted F1, both are greatly affected by class-prevalence. As discussed in \S \ref{subsec:wf1}, we could not determine clear reasons for favoring weighted F1 over Accuracy.

\subsection{What are other `balances'?} 

The concept of `balance' seems positively flavored, and thus we may wish to reflect on more `balances' other than prevalence invariance (\propref{pi}).

Another type of `balance' is introduced by $GmacR$ (or $HmacR$). By virtue of the geometric (harmonic) mean that puts more weight on low outliers, they favor a classifier that equally distributes its correctness over classes. This is also reflected by $macR$ being \textit{strictly} chance corrected with $1/n$, while its siblings have $1/n$ as the upper bound \textit{only} achieved by the \textit{uniform random baseline}, and the metrics' gradients that scale with low-recall outliers (Appendix \ref{app:ccmacr}, \ref{app:gradientghmacr}). 

Yet again another type of `balance' we saw in macro F1 ($macF1$) that selects a classifier with high recall over many classes (as featured by a `macro' metric) and maximizes empirical data set correctness (as featured by a `micro' metric), an attribute that is also visible in its chance baseline upper bound ($1/n$) that is \textit{only} achieved by a prevalence-informed baseline.

Finally, a `meta balance' could be achieved when we are unsure which metric to use, by ensembling a score from a set of selected metrics.

\subsection{Value of class-wise Recall}

Interestingly, from class-wise recall scores we can guess the precision scores in another data set (in the absence of a reference). First, we state an estimate of the class distribution $\mathcal{P}(c)\approx\widehat{\mathcal{P}(c)}$ that can be expected. Then we estimate $\mathcal{P}(f)\approx\widehat{\mathcal{P}(f)}$, simply by running the classifier. Finally, the precision for a class $i$ will then equal
\begin{align*}
    \frac{\mathcal{P}(f \rightarrow i | c=i)\cdot{\mathcal{P}(c=i)}}{\mathcal{P}(f \rightarrow i)} \approx \frac{R_i \cdot \reallywidehat{\mathcal{P}(c=i)}}{\reallywidehat{\mathcal{P}(f \rightarrow i)}}.
\end{align*}
Estimated scores of macro metrics follow. It is not possible to project new recall values from old precision scores (since these do not transfer), underlining the value of recall statistics. Note that this is an idealized approximation, and complex phenomena such as domain shifts are (same as in other parts of this work) not at all accounted for.

\begin{table*}
\centering
\scalebox{0.58}{
\begin{tabular}{ll|r|rrrrrrrrrrrrr|rrrrrrrr}
\toprule
    
& & & \multicolumn{13}{c}{standard} & \multicolumn{4}{c}{after prevalence calibration}\\
sys & off.\ r & citations &Acc & macR & macP & macF1 & macF1' & weightF1 & Kappa & MCC & GmacR & HmacR & r1 & r2 & r3 & macF1 & macF1' & Kappa & MCC \\ 
\midrule
A & 1 & 555 & 65.1 & \textbf{68.1} & 65.5 & 65.4 & 66.8 & 64.5 & 46.5 & 48.0 & 66.8 & 65.4 & 82.9 & 51.2 & \textbf{70.2} & \textbf{67.7} & 68.2 & 52.1 & 52.6 \\
B & 1 & 271 &  65.8 & \textbf{68.1} & \textbf{67.4} & 66.0 & \textbf{67.8} & 65.1 & \textbf{47.3} & \textbf{49.2} & 66.5 & 65.0 & \textbf{87.8} & 51.4 & 65.2 & \textbf{67.7} & \textbf{68.7} & \textbf{52.2} & \textbf{53.1} \\
C & 3 & 20 &  66.1 & 67.6 & 66.2 & 66.0 & 66.9 & 65.7 & 47.2 & 48.1 & 66.8 & 66.0 & 81.7 & 56.0 & 65.2 & 67.5 & 68.0 & 51.4 & 51.8 \\
D & 4 & 12 & 65.2 & 67.4 & 64.9 & 65.1 & 66.1 & 64.8 & 46.3 & 47.3 & 66.5 & 65.6 & 80.3 & 54.2 & 67.6 & 67.1 & 67.5 & 51.0 & 51.3 \\ 
E & 5 & 23 &\textbf{ 66.4} & 66.9 & 65.4 & \textbf{66.0} & 66.1 & \textbf{66.4} & 47.0 & 47.0 & \textbf{66.8} & \textbf{66.8} & 69.8 & \textbf{64.0} & 66.8 & 67.3 & 67.5 & 50.3 & 50.5  \\
F & 6 &11 & 64.8 & 65.9 & 63.9 & 64.5 & 64.9 & 64.7 & 45.0 & 45.4 & 65.6 & 65.4 & 73.5 & 58.7 & 65.6 & 66.1 & 66.3 & 48.9 & 49.0 \\
G & 7 & 2 & 63.3 & 64.9 & 63.6 & 63.4 & 64.2 & 63.1 & 43.0 & 43.8 & 64.2 & 63.5 & 77.4 & 53.9 & 63.5 & 64.9 & 65.4 & 47.4 & 47.7 \\
H & 8 & 30 & 64.3 & 64.5 & 63.1 & 63.7 & 63.8 & 64.4 & 43.6 & 43.6 & 64.5 & 64.5 & 65.3 & 63.6 & 64.5 & 64.9 & 65.2 & 46.7 & 47.0 \\
\bottomrule
\end{tabular}}
\caption{Shared task ranking. `off.\ r' is the official rank of a system. r$i$: recall for class $i$.}
\label{tab:ranking}
\end{table*}

\section{Reflecting on SemEval shared tasks}
 \label{sec:semeval}

So far, we had focused on theory. Now we want to take a look at applied evaluation practice. We study works from the \textit{SemEval} series, a large annual NLP event where teams compete in various tasks, many of which are classification tasks.

\subsection{Example shared task study}
 
As an example, we first study the popular SemEval shared task \cite{rosenthal2017semeval} on tweet sentiment classification (positive/negative/neutral) with team predictions thankfully made available. 
  
\paragraph{Insight: Different metric $\rightarrow$ different ranking.} In Table \ref{tab:ranking} we see the results of the eight best of 37 systems.\footnote{For descriptions of systems A...H, see: \citet{baziotis-etal-2017-datastories-semeval-A,cliche-2017-bb-B,rouvier-2017-lia-C,hamdan-2017-senti17-D,yin-etal-2017-nnembs-E,kolovou-etal-2017-tweester-F,miranda-jimenez-etal-2017-ingeotec-G,jabreel-moreno-2017-sitaka-H}.} The two `winning' systems (A, B, Table \ref{tab:ranking}) were determined with $macR$, which is legitimate. Yet, system E also does quite well: it obtains highest Accuracy, and it achieves a better balance over the three classes ($R_1=69.8, R_2 = 64.0, R_3=66.8$, max.\ $\Delta=5.8$) as opposed to, e.g., system B  ($R_1=87.8, R_2=51.4$, $R_3=65.2$ max.\ $\Delta=36.4$), indicated, on average, also by $GmacR$ and $HmacR$ metrics. So if we want to ensure performance under high uncertainty of class prevalence (as expected in Twitter data?), we may prefer system E, a system that would be also ranked higher when using an ensemble of metrics. 

\begin{figure}
    \centering
    \includegraphics[width=1.0\linewidth]{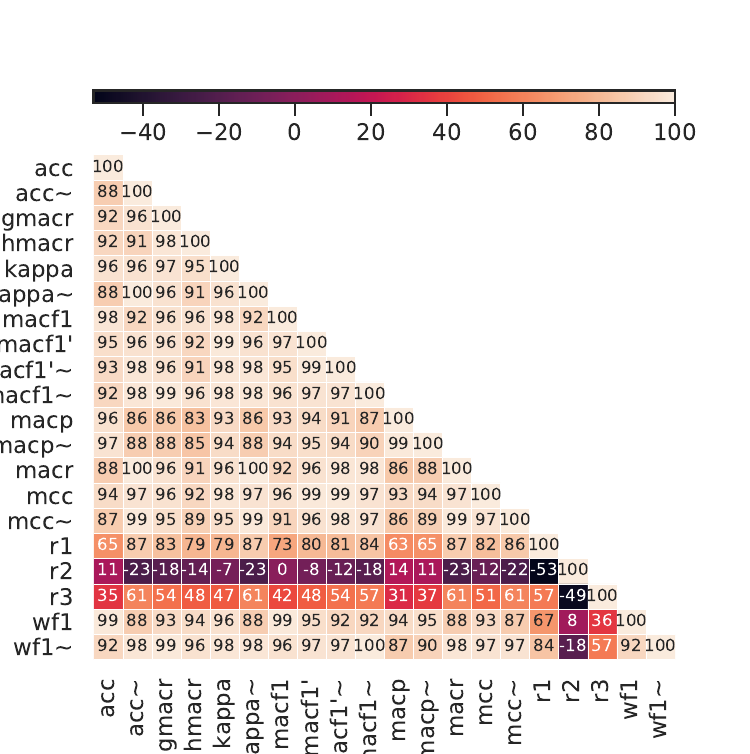}
    \caption{Team ranking correlation matrix wrt.\ metrics. \textit{metric}\char`\~~ means that the confusion matrix has been calibrated before \textit{metric} computation (Eq. \ref{eq:calibrate}).}
    \label{fig:rank}
\end{figure}

Figure \ref{fig:rank} shows a pair-wise Spearman $\rho$ correlation of team rankings of all 37 teams according to different metrics. There are only 4 pairs of metrics that are in complete agreement ($\rho=100$): Macro Recall agrees with calibrated Kappa and calibrated Accuracy. This makes sense: As we before (Eq.\ \ref{eq:macacc=macrSTART}, Eq.\ \ref{eq:kappa_calib}), they are equivalent (the same applies to weighted F1 and macro F1, after calibration).  Looking at single classes, it seems that the second class is the one that can tip the scale: $R_2$ disagrees in its team ranking with \textit{all} other metrics ($\rho \leq 14$). 

\paragraph{Do rankings impact paper popularity?} For the eight best systems in Table \ref{tab:ranking}, we retrieve their citation count from Google scholar as a (very coarse) proxy for their popularity. The result invites the speculation that a superficial ordering could already be effectual: While both team 1 (A) and team 2 (B) were explicitly selected as the winner of the shared task, the first-listed system appears to have almost double the amount of citations (even though B performs better as per most metrics). The citations of other systems (C--H) do not exceed lower double digits, although we saw that a case can even be made for E (rank 6) that achieves stable performance over all classes.

\subsection{Examining metric argumentation}

Our selected example task used macro Recall for winner selection. While the measure had been selected with care (its value became also evident in our analysis), we saw that systems were very close and arguments could have been made for selecting a slightly different (set of) winner(s). To our surprise, in a large proportion of shared tasks, the situation seems worse. Annotating 42 classification shared task overview papers from the recent 5 years\footnote{A csv-file with our annotations is accessible at \url{https://github.com/flipz357/metric-csv}.}, we find that 
\begin{itemize}
      \item only 23.8\% of papers provide a formula, and
      \item only 10.9\%  provide a sensible argument for their metric. 14.3\% use a weak argument similar to the ones shown in Table \ref{tab:practices}. A large amount (73.8\%) does not state an argument or employ a `trope' like ``As is standard, ...''. 
\end{itemize}
The most frequent metric in our sample is `macro F1' ($macF1$). In some cases, its doppelganger $macF1'$ seems to have been used, or a `balanced Accuracy'\footnote{We did not find a formula. As per \texttt{scikit-learn} (\url{https://scikit-learn.org/stable/modules/generated/sklearn.metrics.balanced_accuracy_score.html}, 2024/02/05) it may equal macro Recall.}. Sometimes, in the absence of further description or formula, it can be hard to tell which `macro F1' metric has been used, also due to deviating naming (macro-average F1, mean F1, macro F1, etc.). In at least one case, this has led to a disadvantage: \citet{fang-etal-2023-epicurus} report that ``\textit{During model training and validation, we were not aware that the challenge organizers used a different method for calculating macro F1, namely by using the averages of precision and recall}'', a measure that is much different (cf.\ \S \ref{subsec:macrof1}). 

We also may wonder about the situation beyond SemEval. While a precise characterization of a broader picture would escape the scope of this paper, a cursory glance shows the observed unclarities in research papers from all kinds of domains.

\section{Recommendations}
  \label{sec:recommendations}
Overall, we would like to refrain from making bold statements as to which metric is `better', since different contexts may easily call for different metrics. Still, from our analyses, we can synthesize some general recommendations:
  \begin{itemize}[itemsep=0em]
      \item State the evaluation metric clearly, best use a formula. This also helps protect against ambiguity, e.g., as induced through homonyms such as $macF1$ and $macF1'$.
      \item Try building a case for a metric: As a starting point we can think about how the class distribution in our data set would align with the distribution that we could expect in an application. With greater uncertainty, more macro metric features may be useful. For finer selection, consider viewing our analyses, checking any desirable metric features. Our summary (\S \ref{sec:summary}) can provide guidance.
      \item Consider presenting more than a single number. In particular, complementary metrics such as Accuracy and macro Recall can be indicative about i) a classifier's empirical data set correctness (as in a micro metric) and ii) its robustness to class distribution shifts (as in a macro metric). If the amount of classes $n$ is low, consider presenting class-wise recall scores for their generalizability. If $n$ is larger and a metric is decomposable over classes (\propref{cd}), reporting the variance of a `macro' metric over classes can also be of value.
      \item Consider admitting multiple `winners' or `best systems': If we are not able to build a strong case for a single metric, then it may be sensible to present a \textit{set of well-motivated metrics} and select a \textit{set of best systems}. If \textit{one} `best' system needs be selected, an average over such a set could be a useful heuristic.
  \end{itemize}

\section{Background and related work}
\label{sec:rw}

\paragraph{Meta studies of classification metrics.} Surveys or dedicated book chapters can provide a useful introduction to classification evaluation \cite{sokolova2009systematic, manning2009introduction, tharwat2020classification, grandini2020metrics}. Deeper analysis has been provided mostly in the two-class setting: In a series of articles by David MW Powers, we find the (previously mentioned) Bookmaker's perspective on metrics \cite{powers2020evaluation}, a critique of the F1 \cite{powers2015f} and Kappa \cite{powers-2012-problem}. \citet{10.1371/journal.pone.0222916_kappa_mcc} study binary MCC and Kappa (favoring MCC), while \citet{fabrizio} defines axioms for binary evaluation, including a monotonicity axiom akin to a stricter version of our \propref{mo}, advocating a `K-metric'.\footnote{Same as \citet{powers2020evaluation}'s \textit{Informedness}, the \textit{K-metric} can be understood as two-class macro Recall.} \citet{luque2019impact} analyze binary confusion matrices, and \citet{chicco2020advantages} compare F1, Accuracy and MCC, concluding that ``MCC should be preferred (...) by all scientific communities''. The mathematical relationship between the two macro F1s is further analyzed by \citet{opitz2019macro}. 

Overall, we want to advocate for a mostly \textit{agnostic stance} as to what metric might be picked in a case (if it is sensibly done so), remembering our premise from the introduction: \textit{the perfect metric doesn't exist}. Thus, we aimed at balancing intuitive interpretation and analysis of metrics, while acknowledging desiderata as worded in papers.

\paragraph{Other classification evaluation methods} can be required when class labels reside on a nominal `Likert' scale \cite{o2017overview, amigo-etal-2020-effectiveness}, or in a hierarchy \cite{kosmopoulos2015evaluation}, or they are ambiguous and need be matched (e.g., `none'/`null'/`other') across annotation styles \cite{fu2020interpretable}, or their number is unknown \cite{ge2017generative}. Classifiers are also evaluated with \textit{P-R curves} \cite{NIPS2015_33e8075e} or a \textit{receiver-operator characteristic} \cite{fawcett2006introduction, honovich-etal-2022-true-evaluating}. The \textit{CheckList} \cite{ribeiro-etal-2020-beyond} proposes behavioral testing of classifiers, and the \textit{NEATCLasS} workshop series \cite{neatclass} is an effort to find novel ways of evaluation.

\section{Conclusion}
\label{sec:remarks}

Starting from a definition of the two basic and intuitive concepts of \textit{classifier bias} and \textit{class prevalence}, we examined common classification evaluation metrics, resolving unclear expectations such as those that pursue a `balance' through `macro' metrics. Our metric analysis framework, including definitions and properties, can aid in the study of other or new metrics. A main goal of our work is to provide guidance for more informed decision making in metric selection.

\section*{Acknowledgements}

We are grateful to three anonymous reviewers and Action Editor Sebastian Gehrmann for their valuable comments. We are also thankful to Julius Steen for helpful discussion and feedback.

\bibliography{tacl2021}
\bibliographystyle{acl_natbib}

\appendix

\section{Accuracy aka micro Precision/Recall/F1}
\label{app:micacc}

Micro F1 is defined\footnote{E.g., c.f., \citet{sokolova2009systematic}.} as the harmonic mean ($HM$) of `micro Precision' and `micro Recall', where micro Precision and micro Recall are 
\begin{equation*}
\frac{\sum_x correct(x)}{\sum_x bias(x)}~~\text{and}~~\frac{\sum_x correct(x)}{\sum_x prevalence(x)}.
\end{equation*}
Now it suffices to see that $\sum_x prevalence(x)=\sum_x bias(x)$ and $HM(a,a)=a$.

\subsection{Monotonicity $\checkmark$}
$i$$\neq$$j$: $\frac{\partial Acc(m)}{\partial m_{i,j}}=-\frac{\sum_k correct(k)}{|S|^2} = -\frac{Acc}{|S|}\leq 0$; else $\frac{\partial Acc(m)}{\partial m_{i,i}}=\frac{|S| - \sum_k correct(k)}{|S|^2} = \frac{1- Acc}{|S|} \geq 0$.

\subsection{Other properties}

It is easy to see that properties other than Monotonicity are not featured by Accuracy. 

\section{Macro Recall}
\label{app:anamacr}
\subsection{Monotonicity \cmark}

If $i\neq j$: $\frac{\partial macR(m)}{\partial m_{i,j}}=-\frac{correct(j)}{n \cdot prevalence(j)^2} \leq 0$; else $\frac{\partial macR(m)}{\partial m_{i,i}}=\frac{prevalence(i)- correct(i)}{n \cdot prevalence(i)^2} \geq 0.$

\subsection{Class sensitivity \cmark}

Follows from above.

\subsection{Class decomposability \cmark}

In Eq.\ \ref{eq:mean} set  $g(row, col, x) = \frac{row_x}{\sum_i col_i}$ and $p=1$.

\subsection{Prevalence invariance \cmark}

$R'_i = \frac{\lambda_{i,i}m_{i,i}}{\sum_{j}\lambda_{i,i} m_{j,i}} =\frac{\lambda_{i,i}m_{i,i}}{\lambda_{i,i}\sum_{j} m_{j,i}} = R_i$.

\subsection{Chance correction \cmark}
\label{app:ccmacr}

Assume normalized class prevalences $p \in [0,1]^n ~\text{s.t.}~ \sum_{i=1}^n p_i=1$ and arbitrary random baseline $z \in [0,1]^n ~\text{s.t.}~ \sum_{i=1}^n z_i=1$:
\begin{align*}
    MacR &= \frac{1}{n} \sum_i R_i = \frac{1}{n} \sum_i \frac{p_i \cdot z_i}{p_i} = \frac{1}{n},\\
    GmacR &= \sqrt[n]{\prod_{i=1}^n z_i} \leq \frac{1}{n},\\
    HmacR &= \frac{n}{\sum_{i=1}^n \frac{1}{z_i}} \leq \frac{n}{\sum_{i=1}^n n} = \frac{1}{n}.
\end{align*}
We see that all macro Recall variants are chance corrected. $macR$ is strictly chance corrected.

\subsection{Gradients for $GmacR$ and $HmacR$}
\label{app:gradientghmacr}
For comparison we also include $macR$ with arithmetic mean ($AM$). Let $\mu_x = (n \cdot prevalence(x))^{-1}$, then ($i,j$ implies $i \neq j$):
\begin{align*}
    &\frac{\partial AM}{\partial m_{i,i}} = \mu_i (1 - R_i);~~~~~~~~~~~~~\frac{\partial AM}{\partial m_{i,j}} = -\mu_j R_j;\\
    &\frac{\partial GM}{\partial m_{i,i}} = GM\mu_i (R_i^{-1} - 1);~~\frac{\partial GM}{\partial m_{i,j}} = -GM\mu_j;\\
    &\frac{\partial HM}{\partial m_{i,i}} = HM^{2}\mu_i (R_i^{-2} - R_i^{-1});\\
    &\frac{\partial HM}{\partial m_{i,j}} = -HM^2\mu_jR_j^{-1}.~~~~~~~~~~~~\\
\end{align*}

\section{Macro Precision}
\label{app:anamacp}

\subsection{Monotonicity \cmark}

If $i\neq j$: $\frac{\partial macP(m)}{\partial m_{i,j}}=-\frac{correct(i)}{n \cdot bias(i)^2} \leq 0$; else $\frac{\partial macP(m)}{\partial m_{i,i}}=\frac{bias(i)- correct(i)}{n \cdot bias(i)^2} \geq 0 $.

\subsection{Class sensitivity \cmark}

Follows from above.

\subsection{Class decomposability \cmark}

In Eq.\ \ref{eq:mean} set $g(row, col, x) = \frac{row_x}{\sum_i row_i}$ and $p=1$.

\subsection{Prevalence invariance}
\label{app:macp-pi}

A counter-example $P'_i= \frac{\lambda_{i,i} m_{i,i}}{\sum_{j}\lambda_{j,j}m_{i,j}} \neq P_i$ is easily found. E.g., in Table \ref{tab:subtab1}, \ref{tab:subtab2}, \ref{tab:subtab3}: $macP = 0.5\frac{3}{4} + 0.5\frac{1}{2} = \frac{5}{8} \neq macP' = 0.5\frac{3}{5} + 0.5\frac{2}{3} =\frac{19}{30}$.

\subsection{Chance correction \cmark}
\label{app:ccmacp}

Given same assumptions as in \ref{app:ccmacr} we have
\begin{equation*}
    \frac{1}{n} \sum_i P_i = \frac{1}{n} \sum_i \frac{p_i \cdot z_i}{\sum_j z_i \cdot p_j} = \frac{1}{n} \sum_i p_i = \frac{1}{n}.
\end{equation*}

\section{Macro F1}
\label{app:anamacf1}

\subsection{Monotonicity \cmark}

Let $Z_x=bias(x) + prevalence(x)$. If $i\neq j$: 
\begin{equation*}
    \frac{\partial macF1(m)}{\partial m_{i,j}} = -\frac{2correct(i)}{nZ_i^2} - \frac{2correct(j)}{nZ_j^2} \leq 0
\end{equation*}
else:
\begin{align*}
    \frac{\partial macF1(m)}{\partial m_{i,i}} &= \frac{2}{nZ_i} - \frac{2 \cdot correct(i)}{nZ_i^2} \geq 0.
\end{align*}

\subsection{Class sensitivity \cmark}

Follows from above.

\subsection{Class decomposability \cmark}

In Eq.\ \ref{eq:mean} set $p=1$, $g(row, col,x) =\frac{2row_x}{\sum_i row_i + col_i}$.

\subsection{Prevalence invariance}

It is not prevalence-invariant, c.f. \ref{app:macp-pi}.

\subsection{Chance correction \cmark}
\label{app:ccmacf1}
Given same assumptions as in \ref{app:ccmacr} we have
\begin{align}
    MacF1 &= \frac{1}{n} \sum_i \frac{2 \cdot p_i \cdot z_i}{\sum_j z_i \cdot p_j + \sum_j z_j \cdot p_i} \\ &= \frac{1}{n} \sum_i \frac{2 \cdot p_i \cdot z_i}{p_i +z_i}.
\end{align}
We see that a maximum is attained when p = z, and that this maximum is $\frac{1}{n}$. 

\section{Macro F1 (name twin)}
\label{app:anamacf1bar}

\subsection{Monotonicity \cmark}

Let $(macR + macP - macP \cdot macR) = \epsilon \geq 0$. We have $\frac{\partial macF1'(m)}{\partial m_{i,j}} = \frac{2x\epsilon}{macR + macP}$ where $x=(\frac{\partial macR(m)}{\partial m_{i,j}} + \frac{\partial macP(m)}{\partial m_{i,j}})$. Since $macR$ and $macP$ are monotonic, $macF1'$ also has monotonicity.

\subsection{Label sensitivity \cmark}

Follows from above.

\subsection{Class decomposability}

Not possible.

\subsection{Prevalence invariance}

It is not prevalence-invariant, c.f. \ref{app:macp-pi}.

\subsection{Chance correction \cmark}
\label{app:ccmacf1bar}
Since $macF1'$ is the $HM$ of(strictly chance corrected) macro Precision and macro Recall, we also have strictly chance correction with $\frac{1}{n}$. 

\section{Weighted F1}
\label{app:wf1}

\subsection{Monotonicity}

For brevity, let $prevalence(i)$, $bias(i)$, $correct(i)$ $=$ $x_i$, $y_i$, $z_i$. If $i\neq j$:
\begin{align*}
   & \frac{\partial weightF1(m)}{\partial m_{i,j}} =\\
   &\frac{1}{|S|}\bigg( \frac{2y_jz_j}{(x_j + y_j)^2} - \frac{2x_iz_i}{(x_i + y_i)^2} - weightF1\bigg) \overset{\text{\large \lightning}}{\leq} 0. 
\end{align*}
If we fix the positive term, and let the others approach zero, then is a situation, where $weightF1$ increases even though a classifier made an error.

\subsection{Label sensitivity  \cmark}

Follows from above.

\subsection{Class decomposability}

Not possible.

\subsection{Prevalence invariance}

Trivial.

\subsection{Chance correction}

Trivial.

\section{Kappa and MCC}
\label{app:anamcckappa}
\subsection{Monotonicity}
\label{app:proof1}

We resort back to Kappa and MCC formulas for non-normalized confusion matrices, multiplying numerator and denominator by $s^2$, where $s= \sum_{(i,j)} m_{i, j}$ is the number of data examples, and write $r$ for $\sum_i{correct(i)}$.

\paragraph{Kappa.} We have
\begin{equation}
    KAPPA = \frac{rs - \mathbf{p^Tb}}{s^2 - \mathbf{p^Tb}} = \frac{N_{K}}{D_K}.
\end{equation}
Other variables were introduced before (Eq.\ \ref{eq:kappa_mcc_helper_def}). Now, let $z_{ij} = prevalence(i) + bias(j)$. 

Then,  iff $i \neq j $:
\begin{equation}
   \frac{\partial KAPPA}{\partial m_{i,j}} =  \frac{(r - z_{ij})D_K^2 - (2s -z_{ij})N_K}{D_{K}^2}.
\end{equation}

\paragraph{MCC.} Let us state
\begin{equation}
    MCC = \frac{rs - \mathbf{p^Tb}}{\sqrt{(s^2 - \mathbf{p^Tp})(s^2 - \mathbf{b^Tb})}} = \frac{N_{M}}{D_M}
\end{equation}
Let now $v_{ij} = \frac{\partial p^Tp}{\partial m_{i,j}} = 2\cdot prevalence(j)$ and  $u_{ij} = \frac{\partial b^Tb}{\partial m_{i,j}} = 2\cdot bias(i)$.

Then, iff $i \neq j$:
\begin{align*}
   \frac{\partial MCC}{\partial m_{i,j}} = \frac{1}{2D_{M}^3}\bigg[&D_M^2(r - z_{ij}) \\
    &- N_M(2s - v_{ij})\sqrt{s^2 - \mathbf{b^Tb}} \\
    & - N_M(2s - u_{ij})\sqrt{s^2 - \mathbf{p^Tp}} \bigg].
\end{align*}
It suffices now to see that there exist configurations of confusion matrices where $N_K$ (Kappa) or $N_M$ (MCC) $\rightarrow$ 0, but not $(r - z_{i,j})\cdot{D_{M|K}^2} \rightarrow 0$. \qed

An example, where MCC increases, when we add more errors, is described in Tables \ref{tab:subtabx1} and \ref{tab:subtabx2}.

\begin{table}
\begin{minipage}{.5\linewidth}
    \centering
    \begin{tabular}{cc|ccc} 
    & & \multicolumn{3}{c}{$c =$}\\
          &  & x & y & z \\
        \hline
         & x   & 10  & 43 & 0 \\
        &y   & 1 & 1 & 0\\ 
          \rot{\rlap{$~~f\rightarrow$}}
        &z  & 0 & 0 & 1 \\
    \end{tabular}
    \caption{\\MCC = \textbf{0.0}\\
             Kappa = \textbf{0.0}.}
    \label{tab:subtabx1}
\end{minipage}%
\begin{minipage}{.5\linewidth}
    \centering
\begin{tabular}{cc|ccc} 
    & & \multicolumn{3}{c}{$c =$}\\
     & & x& y & z \\
        \hline
        & x   & 10  & 43 & 0 \\
        & y   & 1 & 1 & 0\\
                  \rot{\rlap{$~~f\rightarrow$}}
        & z  & 0 & \textbf{10} & 1 \\
    \end{tabular}
    \caption{\\MCC = \textbf{0.07} \\
    Kappa = \textbf{0.02}.}
    \label{tab:subtabx2}
\end{minipage}
\label{tab:exx1}
\end{table}

\subsection{Class sensitivity \cmark}
 Trivial.

\subsection{Class decomposability}
Trivial.

\subsection{Prevalence invariance}
Trivial.

\subsection{Chance correction \cmark}
\label{app:cckmcc}
Given same assumptions as in \ref{app:ccmacr}, in the numerators we have
\begin{equation*}
    \sum_i p_i \cdot  z_i - \sum_i \bigg(\sum_j  z_i \cdot p_j\bigg)\bigg(\sum_j z_j \cdot p_i\bigg) = 0.
\end{equation*}

\end{document}